\documentclass[letterpaper]{article} 
\usepackage{aaai25}  
\usepackage{times}  
\usepackage{helvet}  
\usepackage{courier}  
\usepackage[hyphens]{url}  
\usepackage{graphicx} 
\usepackage{natbib}  
\usepackage{caption} 
\usepackage{algorithm}
\usepackage{algorithmic}
\usepackage{amsfonts}
\usepackage{amsmath}
\usepackage{booktabs}
\usepackage{newfloat}
\usepackage{listings}
\DeclareCaptionStyle{ruled}{labelfont=normalfont,labelsep=colon,strut=off} 
\floatstyle{ruled}
\newfloat{listing}{tb}{lst}{}
\floatname{listing}{Listing}
\title{Riemannian Geometric-based Meta Learning}
\author {
    JuneYoung Park\textsuperscript{\rm 1,2}\equalcontrib, 
    YuMi Lee\textsuperscript{\rm 3}\equalcontrib,\
    Tae-Joon Kim\textsuperscript{\rm 2}\thanks{Corresponding Author.},\
    Jang-Hwan Choi\textsuperscript{\rm 3}\footnotemark[2]
}
\affiliations {
    \textsuperscript{\rm 1}Opt-AI Inc. \\
    \textsuperscript{\rm 2}Ajou University School of Medicine\\
    \textsuperscript{\rm 3}Ewha Womans University \\
    jyoung.park@opt-ai.kr, eumi@ewha.ac.kr, tjkim23@ajou.ac.kr, choij@ewha.ac.kr
}

\begin{document}

\maketitle

\begin{abstract}
Meta-learning, or ``learning to learn," aims to enable models to quickly adapt to new tasks with minimal data. While traditional methods like Model-Agnostic Meta-Learning (MAML) optimize parameters in Euclidean space, they often struggle to capture complex learning dynamics, particularly in few-shot learning scenarios. To address this limitation, we propose Stiefel-MAML, which integrates Riemannian geometry by optimizing within the Stiefel manifold, a space that naturally enforces orthogonality constraints. By leveraging the geometric structure of the Stiefel manifold, we improve parameter expressiveness and enable more efficient optimization through Riemannian gradient calculations and retraction operations. We also introduce a novel kernel-based loss function defined on the Stiefel manifold, further enhancing the model’s ability to explore the parameter space. Experimental results on benchmark datasets—including Omniglot, Mini-ImageNet, FC-100, and CUB—demonstrate that Stiefel-MAML consistently outperforms traditional MAML, achieving superior performance across various few-shot learning tasks. Our findings highlight the potential of Riemannian geometry to enhance meta-learning, paving the way for future research on optimizing over different geometric structures.
\end{abstract}

\section{Introduction}

Meta-learning is a framework designed to enhance the ability of models to rapidly adapt to new tasks by leveraging prior knowledge gained from diverse learning experiences~\cite{vilalta2002perspective, vanschoren2018meta}. This approach is particularly effective in fields where data acquisition is limited, such as healthcare, natural language processing, robotics, recommendation systems, and finance~\cite{rafiei2023meta, hospedales2021meta, vettoruzzo2024advances}. By enabling models to generalize well from only a few examples, meta-learning offers a promising alternative to traditional deep learning methods, which typically require large amounts of data to perform well on new tasks.

One of the most widely used algorithms in meta-learning is Model-Agnostic Meta-Learning (MAML)~\cite{finn2017model}, which focuses on finding model initialization parameters that allow for fast adaptation to new tasks with minimal updates. MAML operates through an inner loop of task-specific learning and an outer loop that updates the model based on its performance across multiple tasks. While effective, MAML's reliance on parameter optimization in Euclidean space can limit its ability to capture the complex learning dynamics that arise as models scale in size and complexity~\cite{rajeswaran2019meta, antoniou2019train}.

In this paper, we propose Stiefel-MAML, which performs loss computations within a Riemannian manifold to more effectively reflect the characteristics of the loss surface in meta-learning, thereby enabling rapid adaptation. Riemannian manifold operations can better capture the geometric properties of data and models compared to Euclidean space~\cite{boumal2014manopt, amari1998natural}. For instance, in Euclidean space, maintaining the orthogonality of a rotation matrix requires post-optimization procedures like Gram-Schmidt orthogonalization or projection operations to form a symplectic matrix, which can distort the optimization path~\cite{mishra2014fixed}. However, when optimization is performed on a Riemannian manifold, the geometric properties of the manifold automatically satisfy these constraints, allowing the optimization path to reflect the true geometric structure without distortion. Additionally, in tasks like low-rank matrix completion, Riemannian optimization techniques can more effectively incorporate geometric properties~\cite{vandereycken2013low}.

Research by Liu and Boumal~\cite{liu2020simple} further suggests that Riemannian optimization is more efficient and accurate in handling high-dimensional problems compared to methods used in Euclidean space. For example, applying Riemannian geometry to high-dimensional data analysis tasks such as minimum balanced cut, non-negative PCA, and K-means clustering improves both computational speed and accuracy. Similarly, Smith~\cite{smith2014optimization} demonstrates that Newton's method and conjugate gradient methods on Riemannian manifolds achieve faster and more efficient convergence compared to their Euclidean counterparts.

\begin{figure}[t]
\centering
\includegraphics[width=0.4\textwidth]{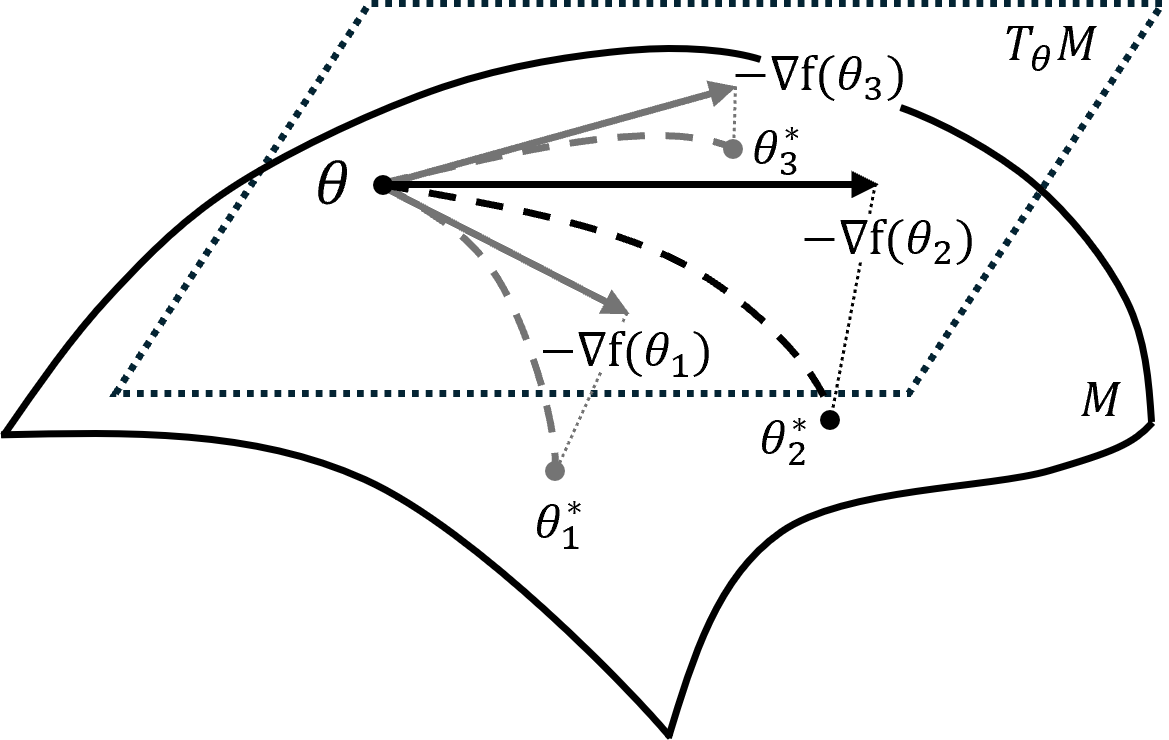} 
\caption{Diagram of the Stiefel-MAML algorithm, which the adaptation of the traditional MAML method within a Riemannian manifold, enabling more efficient navigation and optimization on fixed-manifold (\textit{M}).}
\label{fig:1}
\end{figure}

As illustrated in Figure~\ref{fig:1}, Stiefel-MAML introduces Riemannian geometry into the optimization loop, which allows the model to more effectively capture the geometric properties of the parameter space. By shifting the optimization process onto the Stiefel manifold—a space defined by orthogonal matrices—Stiefel-MAML naturally preserves the underlying geometric constraints throughout the learning process. This inherent preservation of structure, unlike previous studies that applied different manifolds to each task, facilitates effective optimization, even when operating on fixed-manifolds (\textit{M}).
The workflow of Stiefel-MAML follows two main loops. The inner loop performs Riemannian gradient descent on task-specific data, updating the model parameters while preserving the orthogonality constraints imposed by the Stiefel manifold. This is achieved through the use of a retraction operation that projects the updated parameters back onto the manifold after each gradient step. In the outer loop, the meta-parameters are updated by computing the meta-gradient over all tasks. This approach allows the model to adapt quickly to new tasks, capturing the curvature of the parameter space and preventing the optimization path from being distorted.

Through experiments conducted on four benchmark datasets—Omniglot, Mini-ImageNet, FC-100, and CUB—we demonstrate that Stiefel-MAML consistently outperforms conventional MAML in terms of accuracy across all tasks in few-shot learning scenarios. The results indicate that leveraging Riemannian manifolds enables the model to better capture the geometric characteristics of the parameter space, guided by the curvature of the manifold, facilitating faster and more effective adaptation to new tasks.

In summary, the key contributions of this study are as follows:
\begin{itemize}
\item We propose Stiefel-MAML, a novel meta-learning algorithm that leverages Riemannian geometry to improve adaptation in few-shot learning tasks by optimizing model parameters on the Stiefel manifold.
\item We introduce a new kernel-based loss function defined on the Stiefel manifold, which enhances optimization efficiency and task adaptability by capturing the geometric structure of the parameter space.
\item We validate the effectiveness of Stiefel-MAML through extensive experiments on multiple benchmark datasets, demonstrating significant improvements in both accuracy and generalization over existing meta-learning algorithms.
\item We provide a comprehensive analysis of the computational cost associated with Riemannian optimization, showing that the performance gains are achieved with only a marginal increase in computational overhead.
\end{itemize}

By introducing the Stiefel manifold into the meta-learning framework, this study offers a new perspective on optimizing learning processes in non-Euclidean spaces, demonstrating that geometric constraints can be effectively leveraged to enhance model adaptability and performance.

\section{Related Works}
\subsection{Black-box Adaptation}
Black box adaptation in meta-learning refers to a method where the model learns based solely on the input-output relationships without analyzing the internal learning processes. This approach enhances the model's ability to generalize across different tasks, allowing it to quickly adapt to new tasks without the need for explicit access to the model's internal state or gradient information. This contrasts to models such as Recurrent Neural Networks (RNNs) or Long Short-Term Memory networks (LSTMs), which make predictions based on understanding how updates are processed when new inputs are received~\cite{sherstinsky2020fundamentals}. Such learning methods are analogous to the human ability to classify new tasks accurately even with limited examples, making them particularly valuable in fields where data acquisition is challenging.

MAML~\cite{finn2017model} is a type of optimization-based meta-learning that aims to learn generalized initial parameters enabling the model to quickly adapt to new tasks. It has been demonstrated that even without specific task information, MAML can adapt rapidly to new tasks, showcasing strong performance in few-shot learning scenarios and proving the model's adaptability. However, these learning methods do not explicitly address uncertainty. Bayesian meta-learning was developed to address uncertainty directly, thereby improving model optimization based on prior research. BMAML~\cite{yoon2018bayesian} uses a Bayesian approach to quantify Epistemic Uncertainty and learns a posterior distribution over parameters. PMAML~\cite{finn2018probabilistic} represents parameters as probability distributions through Variational Inference and learns a posterior distribution. By incorporating uncertainty into the learning process, these methods improve generalization performance and maximize adaptability to new tasks, thereby enhancing the overall performance of meta-learning.

Despite the advancements in improving generalization and optimization in meta-learning, challenges remain, particularly in capturing complex learning dynamics as models scale. Therefore, this paper proposes addressing these challenges through the use of Riemannian geometry on Stiefel manifolds.

\subsection{Metric-based Meta-learning in Riemannian Manifold and Hyperbolic}
Metric-based Meta-Learning is a method that facilitates rapid learning of new tasks by relying on the distances between data points. This approach operates by embedding data points from similar classes close to each other in the embedded space, while data from different classes are placed farther apart~\cite{koch2015siamese, satorras2018few, chen2020variational}. Representative examples include Prototypical Networks, Matching Networks, and Relation Networks~\cite{sung2018learning}, which enable effective classification of new classes even with a limited amount of training data. These networks typically operate within Euclidean space, where their training processes have been conducted. To overcome this limitation, recent studies have employed embeddings based on Riemannian geometry, which better reflect more intricate geometric structures.

For instance, Gao et al. introduced a data modeling approach using angle-based sectional curvature and manifolds~\cite{gao2022curvature}. They proposed a system of curvature generation and updating, which allows for the initialization of task-specific curvatures and their dynamic adjustment during the training process. This approach effectively captures the intrinsic geometry of the data, shortens the optimization trajectory, and achieves high performance with fewer optimization steps. Additionally, Li et al.~\cite{li2023geometry} proposed Geometry Flow, a metric learning method based on Riemannian geometry. This method defines the path, or ``Flow," that data traverses, learning the geometric transformations that occur as the data moves along the manifold. The Flow is modeled through Riemannian geometry, and this method has shown superior performance, particularly for high-dimensional manifold-structured data. Lastly, First-Order Riemannian Meta-Learning (FORML)~\cite{tabealhojeh2024forml} is a first-order meta-learning method based on Riemannian geometry designed for learning on Stiefel manifolds. While traditional Hessian-free methods have been used in Euclidean space to avoid the complex computation of Hessians in second-order optimization problems~\cite{martens2010deep, absil2008optimization, boumal2014manopt}, FORML extends this approach to Stiefel manifolds, making it well-suited for meta-learning.

\subsection{Justification for Utilizing the Stiefel Manifold in Meta-Learning}

In this study, we choose the Stiefel manifold for meta-learning due to its ability to strictly maintain orthogonality while efficiently handling matrices of various dimensions~\cite{massart2023coordinate, huang2018building}. This property provides several advantages over other Riemannian manifolds. For example, the Grassmann manifold, which focuses on entire subspaces rather than individual basis vectors, is less effective for learning class-specific weight vectors~\cite{huang2018building}. Similarly, while the Special Orthogonal Group enforces orthogonality, it is restricted to square matrices, limiting its usefulness for the non-square weight matrices commonly encountered in deep learning~\cite{li2019orthogonal}. In contrast, the Stiefel manifold preserves orthogonality across a wide range of matrix shapes, reducing parameter correlations and mitigating overfitting—both of which are critical for rapid adaptation in meta-learning scenarios.

Additionally, optimization on the Stiefel manifold is inherently more efficient in terms of computation and memory usage compared to traditional Euclidean approaches. This increased efficiency provides a more direct and practical way to handle non-smooth optimization challenges, which in previous studies often required complex computational strategies~\cite{tabealhojeh2024forml, wang2022riemannian, hu2024constraint}. By leveraging these strengths, the Stiefel manifold facilitates more effective and scalable solutions for meta-learning tasks.

\section{Method}
\subsection{Conventional Meta-Learning}

The core objective of few-shot meta-learning is to develop a model capable of rapidly adapting to new tasks using only a small amount of data and brief training~\cite{gharoun2024meta, jamal2019task}. To achieve this, the learning model undergoes a meta-learning phase composed of various tasks. Through this process, the model becomes able to swiftly adapt to new tasks with just a few examples or attempts. In meta-learning, individual tasks are treated as single training instances. This framework is designed to enable the model to adapt to a variety of tasks represented by a distribution denoted as \(p(T)\).

In a \(K\)-shot learning environment, the goal is to train the model to quickly adapt to a new task \(T_{i}\), randomly selected from \(p(T)\), using only a small amount of information. Specifically, the model must learn using only \(K\) examples and the evaluation criterion \(\mathcal{L}_{T_{i}}\) for that task. The meta-learning process proceeds as follows: First, a task \(T_{i}\) is randomly chosen from \(p(T)\), and \(K\) training examples are extracted. The model learns using these \(K\) examples and the evaluation criterion \(\mathcal{L}_{T_{i}}\). After learning, new test samples are drawn from \(T_{i}\) to evaluate the model's performance.

Based on the test results, the parameters of model \(f\) are updated, with the test errors from each \(T_{i}\) serving as the learning signal for the overall meta-learning process. Once meta-learning is complete, we assess the model's generalization ability by selecting a new, unseen task from \(p(T)\). The model is given only \(K\) examples from this new task to learn from, and after training, its performance is measured to evaluate the effectiveness of the meta-learning. It is crucial to note that the tasks used for meta-testing are different from those used during meta-learning. This ensures that we are truly testing the model's ability to adapt quickly to new tasks.

\begin{algorithm}[ht]
\caption{Stiefel-MAML Algorithm}
\begin{algorithmic}[1]
\REQUIRE Task distribution $p(\mathcal{T})$, learning rates $\alpha, \beta$, number of inner loop steps $K$
\ENSURE Model parameters $\theta$

\STATE Initialize $\theta$ on Stiefel manifold $St(n, p)$
\FOR{each iteration}
    \STATE Sample batch of tasks $\{\mathcal{T}_i\} \sim p(\mathcal{T})$
    \FOR{each task $\mathcal{T}_i$}
        \STATE Initialize task-specific parameters $\theta_i = \theta$
        \FOR{$k = 1$ to $K$} 
        \STATE{Inner loop: Riemannian gradient descent on $St(n, p)$}
            \STATE Compute loss $\mathcal{L}_{\mathcal{T}_i}(\theta_i)$
            \STATE Compute Riemannian gradient $\text{grad} \mathcal{L}_{\mathcal{T}_i}(\theta_i)$
            \STATE Update parameters: 
            \[
            \theta_i \leftarrow R_{\theta_i}(-\alpha \cdot \text{grad} \mathcal{L}_{\mathcal{T}_i}(\theta_i))
            \]
        \ENDFOR
    \ENDFOR
    \STATE Compute meta-gradient:
    \[
    \nabla_{\theta} \sum_{\mathcal{T}_i} \mathcal{L}_{\mathcal{T}_i}(R_{\theta_i}(-\alpha \cdot \text{grad} \mathcal{L}_{\mathcal{T}_i}(\theta_i)))
    \]
    \STATE Update meta parameters $\theta$:
    \[
    \theta \leftarrow R_{\theta}(-\beta \cdot \nabla_{\theta} \sum_{\mathcal{T}_i} \mathcal{L}_{\mathcal{T}_i}(\theta))
    \]
\ENDFOR
\end{algorithmic}
\label{algorithm:Stiefel-MAML}
\end{algorithm}

\subsection{Stiefel-MAML}
The Stiefel-MAML algorithm is a novel approach developed to overcome the limitations of the existing MAML algorithm. The overall workflow of the algorithm can be found in Algorithm~\ref{algorithm:Stiefel-MAML}. MAML is a widely used technique in meta-learning, aimed at learning model initialization parameters that can quickly adapt to a variety of tasks~\cite{yoon2018bayesian}. However, MAML performs optimization in Euclidean space, which is not ideal for problems involving orthogonality constraints. In such cases, performance may degrade if the parameters fail to maintain these constraints during the optimization process. To address this issue, Stiefel-MAML enhances MAML by utilizing optimization on the Stiefel manifold. The Stiefel manifold, denoted as \(St(n,p)\), represents the space of \( n \times p\) orthogonal matrices. Mathematically, it is defined as:
\begin{equation}
    St(n,p) = {X \in \mathbb{R}^{n\times p} : X^T X = I_p}
\end{equation}
where \(X^T\) is the transpose of matrix \(X\), and \(I_p\) is the \(p \times p\) identity matrix. This manifold is ideal for problems requiring orthogonality constraints, as it naturally enforces these constraints throughout the optimization process. Unlike MAML, which relies on Euclidean optimization, Stiefel-MAML employs Riemannian optimization on the Stiefel manifold, better handling problems with orthogonality constraints.

The learning process of Stiefel-MAML involves two main loops: an inner loop that finds optimal parameters for each task and an outer loop that optimizes parameters across all tasks. In the inner loop, the model parameters \(\theta\) are iteratively updated for each task, providing a common starting point for all tasks. When a task \(T_i\) is sampled from the task distribution \(p(T)\), it comes with its own dataset and learning objective, prompting the model to adjust its parameters to meet these task-specific requirements.

In the inner loop, parameter updates are performed using a Riemannian gradient descent on the Stiefel manifold. This ensures that the parameters are optimized while maintaining orthogonality constraints. Specifically, the parameter \(\theta_i\) is updated to minimize the loss on the dataset for each task \(T_i\). The optimization in the inner loop is carried out over \(K\) steps of Riemannian gradient descent, represented by:
\begin{equation}
    \theta_{k+1} = R_{\theta_{k}}(-\alpha \cdot \text{grad}\mathcal{L}_{T}(\theta_{k}))
\end{equation}
Here, \(\alpha\) represents the learning rate, \(R_{\theta}\) denotes the retraction operation, and \(\text{grad} \mathcal{L}_{T}(\theta)\) signifies the Riemannian gradient. The retraction operation, performed through QR decomposition, maps the parameter updates from the tangent space back onto the Stiefel manifold, ensuring the preservation of the manifold’s structure. The calculation of the Riemannian gradient on the Stiefel manifold involves projecting the gradient from Euclidean space onto the tangent space. The Riemannian gradient is given by the equation:
\begin{equation}
    \scriptstyle\text{grad} \mathcal{L}(X) = (I - XX^T)\nabla \mathcal{L}(X) + X skew(X^T \nabla \mathcal{L}(X))
\end{equation}
In this equation, \(I\) represents the identity matrix, \(\nabla \mathcal{L}(X)\) denotes the Euclidean gradient, and \(skew(A)\) refers to the skew-symmetric part of matrix A. This process ensures stable optimization even in problems that require orthogonality constraints.

The outer loop of Stiefel-MAML aims to minimize the average loss across all tasks. Parameter updates in this loop are performed by calculating the meta-gradient, expressed as: 
\begin{equation}
    \theta \leftarrow R_\theta(-\beta \cdot \text{grad} \mathcal{L}(\theta))
\end{equation}
Here \(\beta\) represents the meta-learning rate, and \(\text{grad} \mathcal{L}(\theta)\) denotes the meta-gradient. The outer loop’s optimization is directed towards minimizing the average loss across various tasks, playing a crucial role in the meta-learning process.

Another important element in Stiefel-MAML is the kernel-based loss function. This loss function utilizes the geodesic distance on the Stiefel manifold to measure the similarity between two points.

\begin{table*}[h]
\centering
\resizebox{\textwidth}{!}{
\begin{tabular}{r|ccc|ccc|ccc|ccc}
\toprule
Dataset & \multicolumn{3}{c|}{Omniglot} & \multicolumn{3}{c|}{Mini-ImageNet} & \multicolumn{3}{c|}{FC-100} & \multicolumn{3}{c}{CUB} \\ 
\midrule
Num-Way & 3 & 5 & 10 & 3 & 5 & 10 & 3 & 5 & 10 & 3 & 5 & 10 \\ 
\midrule
MAML (1) ACC & 99.59 & \textbf{98.70} & 94.80 & \textbf{64.32} & 48.70 & - & 46.32 & 31.28 & 17.85 & 70.21 & 57.33 & 40.98 \\ 
\hfill CI  & \(\pm\)0.31 & \(\pm\)\textbf{0.40} & \(\pm\)0.56 & \(\pm\)\textbf{1.65} & \(\pm\)0.92 & - & \(\pm\)1.45 & \(\pm\)0.97 & \(\pm\)0.66 & \(\pm\)0.94 & \(\pm\)1.15 & \(\pm\)0.97 \\ 
\midrule
FO-MAML (1) ACC& 98.13 & 96.82 & 94.67 & 61.09 & 46.77 & - & 43.21 & 30.25 & 16.95 & 70.24 & 56.86 & 40.55 \\ 
\hfill CI  & \(\pm\)0.55 & \(\pm\)0.59 & \(\pm\)0.48 & \(\pm\)2.02 & \(\pm\)1.48 & - & \(\pm\)1.36 & \(\pm\)0.83 & \(\pm\)0.56 & \(\pm\)1.66 & \(\pm\)1.52 & \(\pm\)1.84 \\
\midrule
S-MAML (1) ACC& \textbf{99.67} & 98.25 & \textbf{98.02} & 63.78 & \textbf{50.49} & - & \textbf{49.07} & \textbf{34.81} & \textbf{20.33} & \textbf{73.11} & \textbf{60.15} & \textbf{43.08} \\ 
\hfill CI  & \(\pm\)\textbf{0.30} & \(\pm\)0.36 & \(\pm\)\textbf{0.33} & \(\pm\)1.89 & \(\pm\)\textbf{1.41} & - & \(\pm\)\textbf{1.56} & \(\pm\)\textbf{1.00} & \(\pm\)\textbf{1.11} & \(\pm\)\textbf{1.21} & \(\pm\)\textbf{0.88} & \(\pm\)\textbf{0.90} \\
\midrule
\midrule
MAML (5) ACC & 99.79 & \textbf{99.90} & 98.64 & 76.31 & 63.11 & - & 56.12 & 42.12 & 26.46 & 83.73 & 74.71 & 59.65 \\ 
\hfill CI  & \(\pm\)0.25 & \(\pm\)\textbf{0.10} & \(\pm\)0.18 & \(\pm\)1.58 & \(\pm\)0.46 & - & \(\pm\)1.54 & \(\pm\)1.07 & \(\pm\)0.58 & \(\pm\)1.37 & \(\pm\)1.59 & \(\pm\)1.08 \\ 
\midrule
FO-MAML (5) ACC& 99.47 & 99.16 & 97.87 & 75.24 & 61.53 & - & 57.00 & 42.12 & 25.78 & 84.04 & 74.85 & 58.87 \\ 
\hfill CI  & \(\pm\)0.35 & \(\pm\)0.18 & \(\pm\)0.23 & \(\pm\)1.78 & \(\pm\)1.34 & - & \(\pm\)1.46 & \(\pm\)1.14 & \(\pm\)0.70 & \(\pm\)1.71 & \(\pm\)1.34 & \(\pm\)1.95 \\
\midrule
S-MAML (5) ACC& \textbf{99.92} & 99.69 & \textbf{98.73} & \textbf{78.71} & \textbf{66.39} & - & \textbf{63.63} & \textbf{48.85} & \textbf{33.81} & \textbf{85.09} & \textbf{75.35} & \textbf{61.27} \\ 
\hfill CI  & \(\pm\)\textbf{0.13} & \(\pm\)0.10 & \(\pm\)\textbf{0.21} & \(\pm\)\textbf{1.37} & \(\pm\)\textbf{1.25} & - & \(\pm\)\textbf{1.41} & \(\pm\)\textbf{1.07} & \(\pm\)\textbf{0.73} & \(\pm\)\textbf{0.92} & \(\pm\)\textbf{1.14} & \(\pm\)\textbf{1.72} \\
\bottomrule
\end{tabular}}
\caption{The results of 1-shot (1) and 5-shot (5) learning for 3-way, 5-way, and 10-way tasks across four datasets (Omniglot, Mini-ImageNet, FC-100, and CUB). The table compares the accuracy (ACC) and 95\% confidence intervals (CI) of MAML, FirstOrder(FO)-MAML, and Stiefel(S)-MAML, demonstrating the consistent performance improvement of S-MAML over the other methods. The 5-way experimental results on the Omniglot and Mini-ImageNet datasets for MAML have been extracted from Finn et al.~\cite{finn2017model}.}
\label{table:GBML_MAIN}
\end{table*}

\paragraph{Theorem 1} Let \textit{\(St(n,p)\)} be the Stiefel manifold of \(n \times p\) orthonormal matrices. We define a positive definite kernel function \(K: St(n,p) \times St(n,p) \to \mathbb{R}^{+}\) and a corresponding loss function \(\mathcal{L}: St(n,p) \times St(n,p) \to \mathbb{R}^{+}\) is defined as: \(\mathcal{L}(X, Y) = 1 - K(X, Y)\)
\\ \\
This kernel-based loss function measures the similarity between two points on the Stiefel manifold and is designed to decrease as similarity increases. This enables effective optimization in problems with orthogonality constraints. Specifically, this loss function plays a crucial role in naturally encoding orthogonality constraints by reflecting the geometric structure of the Stiefel manifold. The ultimate goal of Stiefel-MAML is to learn initial parameters that enable rapid adaptation to new tasks in problems with orthogonality constraints. This approach demonstrates higher generalization performance than the existing MAML algorithm, particularly excelling in problems where orthogonality constraints are important. Through optimization on the Stiefel manifold, which can naturally encode orthogonality constraints, Stiefel-MAML enhances the efficiency of meta-learning and provides a model capable of swiftly adapting to various tasks.

\section{Experimental Setup}
\subsection{Datasets}  
We conducted a series of experiments to evaluate the effectiveness of our proposed framework. These experiments primarily focused on few-shot classification, scenario analysis, and an ablation study. The datasets used for these evaluations include Omniglot, Mini-ImageNet, FC-100, and CUB, all of which are widely recognized in the field of meta-learning. By employing these datasets, we comprehensively assessed the performance of our method across a broad range of tasks.

\subsection{Environments}  
We implemented our methodology using Python 3.8, PyTorch 1.8.1, and the torchmeta library \cite{deleu2019torchmeta}, ensuring a consistent experimental environment. The experiments were conducted using an NVIDIA A6000 (48G) GPU.

\subsection{Hyperparameters}  
Following the approach of Finn et al. (2017), we sampled 60,000 episodes for our experiments. We adopted the 4-convolution architecture described by Vinyals et al. (2016) to implement our parameter update method. The learning rates were set to 0.01 for the inner loop and 0.001 for the outer loop. Additionally, we matched the number of gradient steps in the inner loop to those used in the experiments by Finn et al. (2017).

\subsection{Cross-domain Few-shot Learning}  
The goal of meta-learning is not only to achieve high performance within the trained distribution but also to generalize well to out-of-distribution datasets. To test this, we conducted two cross-domain few-shot learning studies to evaluate whether performance could be maintained. The first scenario involved training on a general dataset and then conducting meta-testing on a specific dataset \((G \rightarrow S)\). The second scenario involved training on a specific dataset and then conducting meta-testing on a general dataset \((S \rightarrow G)\). For these experiments, we used the Mini-ImageNet dataset as the general dataset and the CUB dataset (Wah et al. 2011) as the specific dataset.

\section{Experimental Results}
\subsection{Few-Shot Classification}
Experiments were conducted on Stiefel-MAML, MAML, and FirstOrderMAML across four different datasets. We conducted 1-shot and 5-shot learning for 3-way, 5-way, and 10-way tasks, reporting both accuracy and 95\% confidence intervals (CI), as detailed in Table~\ref{table:GBML_MAIN}. The results demonstrate that Stiefel-MAML consistently outperforms the other methods in most cases.

Compared to MAML, Stiefel-MAML showed an average improvement of:
1) +0.48\% on the Omniglot dataset; 2) +1.73\% on Mini-ImageNet; 3) +5.05\% on FC-100; 4) +1.91\% on the CUB dataset. 
Additionally, the following average improvements were observed for each shot:
1) +1.88\% in 1-shot, and 2) +2.81\% in 5-shot.
Thus, our method has demonstrated its ability to effectively adapt to varying datasets and tasks while maintaining high accuracy.

An effect size analysis using Cohen’s d further supports Stiefel-MAML’s superiority, showing large effect sizes in both the 1-shot (d $=$ 0.83) and 5-shot (d $=$ 0.91) scenarios. Additionally, the 5.05\% improvement on the FC-100 dataset (p $<$ 0.02) demonstrates a statistically significant enhancement.

\begin{table}[]
\centering
\resizebox{.45\textwidth}{!}{%
\begin{tabular}{r|cc|cc}
\toprule
Scenario & \multicolumn{2}{c|}{G$\rightarrow{}$S} & \multicolumn{2}{c}{S$\rightarrow{}$G} \\ 
\midrule
Train $\rightarrow{}$ Test & \multicolumn{2}{c|}{Mini $\rightarrow{}$CUB} & \multicolumn{2}{c}{CUB$\rightarrow{}$Mini} \\ 
\midrule
Num-Shot                   & 1 & 5 & 1 & 5 \\ 
\midrule
MAML (5) ACC  & 47.81 & 63.35 & 28.81 & 39.50  \\
\hfill CI  & \(\pm\)1.34 & \(\pm\)1.23 & \(\pm\)0.72 & \(\pm\)0.74 \\
\midrule
FO-MAML (5) ACC  & 45.72 & 61.13 & 26.95 & 38.75 \\
\hfill CI  & \(\pm\)1.44 & \(\pm\)1.37 & \(\pm\)1.07 & \(\pm\)1.15 \\
\midrule
S-MAML (5) ACC  & \textbf{47.97} & \textbf{65.61} & \textbf{30.94} & \textbf{40.13} \\
\hfill CI  & \(\pm\)\textbf{1.61} & \(\pm\)\textbf{1.30} & \(\pm\)\textbf{1.71} & \(\pm\)\textbf{1.53} \\

\bottomrule
\end{tabular}%
}
\caption{The results of cross-domain few-shot learning, comparing each model (G: General, S: Specific). These results illustrate the flexibility of the algorithm, demonstrating how it remains adaptable and effective regardless of the dataset characteristics.}
\label{table:Scenario_Results}
\end{table}

\subsection{Cross-domain Few-shot Classification}
The traditional machine learning paradigm assumes that training and testing data follow the same statistical patterns. However, in reality, unexpected distribution shifts can occur, leading to a decline in model performance when encountering unfamiliar data. This is known as the Out-Of-Distribution (OOD) problem~\cite{liu2021towards}. Given the diverse and unpredictable nature of real-world problems, it is impossible to cover all possible scenarios with just the training data. Therefore, algorithms with strong OOD generalization capabilities can maintain performance even in unforeseen situations, making them effective in real-world environments.

Our proposed Stiefel-MAML algorithm effectively addresses the OOD generalization problem that traditional machine learning models often struggle with. Although existing models perform well only on test data similar to the training data, our algorithm maintains robust performance even when faced with unexpected distribution shifts. This means that Stiefel-MAML more effectively achieves the core goal of meta-learning, which is to ``learn how to learn."
In the face of diverse and unpredictable real-world challenges, our algorithm demonstrates excellent adaptability, even in situations that it has not encountered during training. To validate this generalization ability, we introduced a new evaluation method called a ``scenario study." This approach assesses the algorithm's domain generalization capability by using data from distributions not seen during the meta-testing phase.

Specifically, we designed two contrasting scenarios: one in which the model is trained on a general dataset and tested on a specific domain dataset, and the other where the reverse is done. We used various way and shot settings in these scenarios to test the flexibility of the algorithm. As shown in Table~\ref{table:Scenario_Results}, we conducted evaluations using the Mini-ImageNet general dataset and the CUB specific domain dataset, focusing on 5-way tasks to validate the performance of our approach. In these results, our Stiefel-MAML outperformed the original MAML algorithm, demonstrating its robustness and high adaptability. Stiefel-MAML proves its ability to learn and generalize effectively even in new environments not encountered during training, making it a more suitable solution for complex and ever-changing real-world problems.

\subsection{Gradient Norm Comparison with Conventional Meta-Learning}
\begin{figure}[]
\centering
\includegraphics[width=0.5\textwidth]{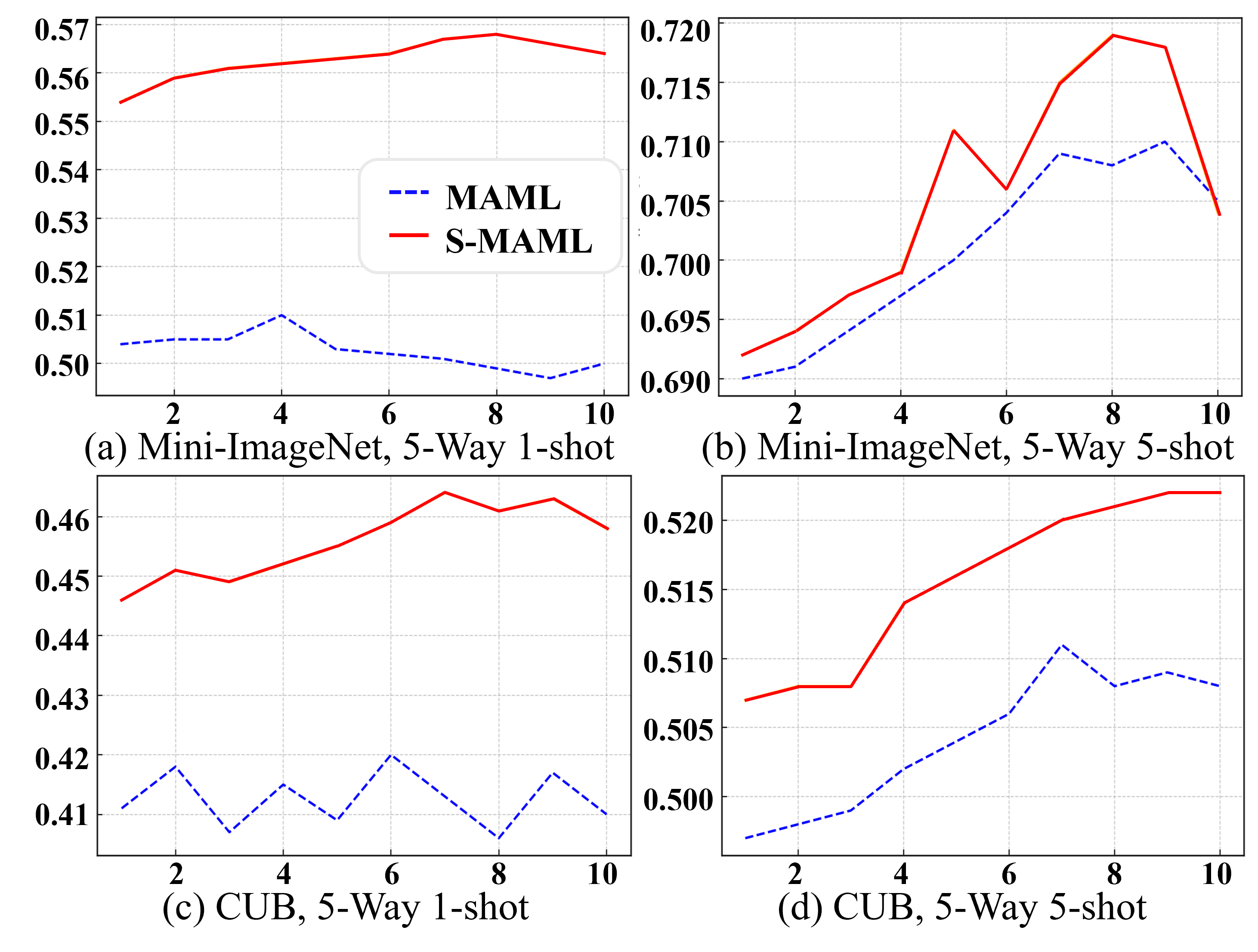}
\caption{Gradient norm results for each task, which the comparative performance between Stiefel(S)-MAML and MAML. Higher gradient norms observed in S-MAML indicate steeper movements along the loss surface, suggesting more effective task adaptation.; x-axis: Adaptation steps; y-axis: Gradient norms}
\label{fig:gradient_norm}
\end{figure}

In this experiment, we compare the magnitudes of the gradient norm between our proposed algorithm, Stiefel-MAML, and the MAML algorithm. The results of the experiment are shown in Figure~\ref{fig:gradient_norm}. First, we evaluate the meta-learned model after performing several adaptation steps on unseen tasks during meta-testing.

The results show that our method consistently exhibits higher gradient norm values than the conventional MAML. This suggests that our algorithm is making steep movements along the loss surface to adapt to the tasks, whereas the conventional algorithms may be approaching local optima, making further improvements challenging. As evidenced by the results in Table~\ref{table:GBML_MAIN}, Stiefel-MAML consistently achieves higher accuracy than the existing algorithms, indicating that our algorithm is more effectively navigating towards the optimum. Furthermore, these results suggest that Riemannian manifolds allow for faster adaptation than Euclidean space.

\begin{table}[t]
\centering
\begin{tabular}{r|ccc}
\toprule
Dataset & \multicolumn{3}{c}{Omniglot} \\ 
\midrule
Num-Way & 3 & 5 & 10\\ 
\midrule
MAML (1) sec/iter & 1.27 & 1.10 & 1.56 \\ 
S-MAML (1) sec/iter & 1.52 & 1.64 & 1.99 \\ 
\midrule
MAML (5) sec/iter & 1.44 & 1.51 & 3.02 \\ 
S-MAML (5) sec/iter & 1.87 & 2.40 & 3.87 \\ 
\bottomrule
\end{tabular}
\caption{Comparison of computational time per iteration, comparing MAML and Stiefel(S)-MAML on the Omniglot dataset across different Num-Way settings. Despite the operations being carried out in Riemannian space, S-MAML does not show a significant increase in computational time compared to MAML.}
\label{table:computational_Results}
\end{table}

\subsection{Computational Cost}
Previous research on neural network algorithms leveraging Riemannian geometry has predominantly utilized symmetric positive definite matrices (SPDs)~\cite{zhao2023modeling}, necessitating complex operations such as matrix exponentials and logarithms. Although these methods effectively capture the characteristics of the loss surface, they exhibit inefficiencies as a result of the need for specific designs tailored to each new manifold. To address this issue, residual networks have been proposed to generalize the geometry of manifolds~\cite{katsman2024riemannian}. However, even these approaches rely on Riemannian exponential maps, resulting in substantial computational demands.

In contrast, our method efficiently captures the properties of Riemannian manifolds without resorting to exponential operations. When evaluated in the Omniglot dataset across different shots, our approach requires slightly more time compared to MAML, but the extent of this increase is not severe (as shown in Table~\ref{table:computational_Results}). Consequently, Stiefel-MAML has proven to be an effective model that ensures high accuracy while demanding less computational capacity. This shows that our approach strikes a balance between computational efficiency and accuracy, making it a viable alternative in scenarios with limited computing resources.

\subsection{Feasibility of Application in Complex Architectures}
Simple CNNs and ResNet differ significantly in their structural characteristics, such as network depth and residual connections, which can lead to variations in learning dynamics. Thus, verifying whether the meta-learning strategy of MAML remains effective in more complex architectures is a critical aspect of evaluation.

To address this, we conducted experiments to assess the performance of Stiefel-MAML compared to MAML on complex architectures like ResNet, rather than simpler CNNs. Using the Omniglot dataset, we evaluated both methods with ResNet50 as the backbone. The results, presented in Table~\ref{table:resnet_Results}, demonstrate that Stiefel-MAML outperforms MAML in the majority of tasks. These findings confirm that the proposed algorithm is not only effective on CNNs but also performs robustly on more complex models like ResNet, highlighting its high applicability across diverse architectures.

\begin{table}[t]
\centering
\begin{tabular}{r|ccc}
\toprule
Model & \multicolumn{3}{c}{ResNet-50} \\ 
\midrule
Num-Way & 3 & 5 & 10\\ 
\midrule
MAML (1) ACC & 93.33 & 90.44 & 72.22 \\ 
\hfill CI  & \(\pm\)0.83 & \(\pm\)0.74 & \(\pm\)1.41 \\
S-MAML (1) ACC & \textbf{94.44} & \textbf{93.33} & \textbf{72.44} \\ 
\hfill CI  & \(\pm\)\textbf{0.40} & \(\pm\)\textbf{0.44} & \(\pm\)\textbf{0.31} \\
\midrule
MAML (5) ACC & \textbf{89.26} & 74.22 & 64.00 \\ 
\hfill CI  & \(\pm\)\textbf{0.36} & \(\pm\)0.71 & \(\pm\)0.61 \\
S-MAML (5) ACC & 86.30 & \textbf{76.22} & \textbf{65.44} \\ 
CI  & \(\pm\)0.42 & \(\pm\)\textbf{0.67} & \(\pm\)\textbf{0.69} \\
\bottomrule
\end{tabular}
\caption{Validation of Stiefel(S)-MAML applicability on ResNet-50 under various Num-Way settings. Unlike simpler CNN architectures, the increased complexity of ResNet-50 does not degrade S-MAML performance, thereby illustrating its robustness.}
\label{table:resnet_Results}
\end{table}

\section{Conclusion}
In this paper, we introduced Stiefel-MAML, an innovative algorithm that transitions parameter updates from Euclidean to non-Euclidean space, leveraging Riemannian geometry in the context of meta-learning. Our approach builds on the premise that non-Euclidean spaces, which inherently incorporate curvature information, can facilitate faster convergence and improve learning dynamics compared to traditional Euclidean spaces. Through extensive experiments, we demonstrated that Stiefel-MAML outperforms conventional algorithms in few-shot learning scenarios, showcasing superior adaptability across diverse datasets in various scenarios. Notably, these improvements are achieved without significant increases in computational cost, ensuring practical efficiency.

While prior research on non-Euclidean spaces has often focused on task-specific optimizations or employed varying manifolds for different tasks, we adopted a fixed-manifold approach using the Stiefel manifold. This contrasts with traditional meta-learning methods, which adapt to specific tasks by altering the underlying manifold. Our findings emphasize the potential of exploring parameter spaces beyond Euclidean geometry, prompting reflection on two fundamental questions: Why confine ourselves to Euclidean space? And must data always be represented as vectors?

By applying this algorithm to meta-learning, we demonstrated its capacity to address the challenges of data scarcity. Meta-learning, with its emphasis on learning and testing with limited data, represents a form of ``data lightening" that is particularly valuable in domains where data is scarce. Our findings underscore the need for further research in these promising directions, expanding the possibilities for meta-learning in real-world applications constrained by limited data.

\section{Acknowledgements}
This work was partly supported by the National Research Foundation of Korea (NRF-2022R1A2C1092072) and by Institute of Information \& communications Technology Planning \& Evaluation (IITP) grant funded by the Korea government (MSIT) (No. RS-2022-00155966, Artificial Intelligence Convergence Innovation Human Resources Development (Ewha Womans University)).

\bibliography{aaai25}

\end{document}